\documentclass[a4paper,alpha-refs]{jaisd}

\usepackage{graphicx}
\usepackage{siunitx}
\usepackage{lipsum}
\usepackage{caption}
\usepackage{lipsum}
\usepackage{xcolor}

\papercat{Position Paper}

\title{From Prediction to Foresight: The Role of AI in Designing Responsible Futures}

\author[1,\authfn{1}]{Mar\'ia P\'erez-Ortiz}

\affil[1]{AI Centre, Department of Computer Science, University College London, London, UK}

\authnote{\authfn{1}maria.perez@ucl.ac.uk}

\runningauthor{P\'erez-Ortiz}

\jvolume{1}
\jnumber{1}
\jyear{2024}
\jstartpage{8}
\jdoi{10.69828/4d4k91}

\begin{document}

\begin{frontmatter}
\maketitle
\begin{abstract}
In an era marked by rapid technological advancements and complex global challenges, responsible foresight has emerged as an essential framework for policymakers aiming to navigate future uncertainties and shape the future. Responsible foresight entails the ethical anticipation of emerging opportunities and risks, with a focus on fostering proactive, sustainable, and accountable future design. \textcolor{black}{This paper coins the term "responsible computational foresight", examining the role of human-centric artificial intelligence and computational modeling in advancing responsible foresight, establishing a set of foundational principles for this new field and presenting a suite of AI-driven foresight tools currently shaping it.}
AI, particularly in conjunction with simulations and scenario analysis, enhances policymakers’ ability to address uncertainty, evaluate risks, and devise strategies geared toward sustainable, resilient futures. However, responsible foresight extends beyond mere technical forecasting; it demands a nuanced understanding of the interdependencies within social, environmental, economic and political systems, alongside a commitment to ethical, long-term decision-making that supports human intelligence. We argue that AI will play a role as a supportive tool in responsible, human-centered foresight, complementing rather than substituting policymaker judgment to enable the proactive shaping of resilient and ethically sound futures. This paper advocates for the thoughtful integration of AI into foresight practices to empower policymakers and communities as they confront the grand challenges of the 21st century.

\medskip


\end{abstract}

\end{frontmatter}

\section{Introduction}

Policymaking is inherently forward-looking. Governments must craft strategies that anticipate and mitigate future risks while seizing emerging opportunities. However, this process is not simply about labeling futures as “good” or “bad.” Rather, the critical question is whether we leave enough room to shape the future \citep{van2019toward}. Predictions often limit this potential by presenting a fixed future, leaving little space for hope or innovation. As \citet{adam2007future} argue, “the future is open, but not empty.” Predictions close off the future rather than opening it up to possibilities and new initiatives. In contrast, responsible foresight \citep{van2019toward} recognizes that while certain challenges of tomorrow—such as climate change or disruptive technologies—may seem daunting, there is still room to shape outcomes through thoughtful, ethical action today.

As technological advancements—particularly in artificial intelligence (AI)—open new possibilities for forecasting and decision-making, speculation grows over whether AI could one day be used as supportive tools for human policymakers. 
This question gains urgency as AI tools, like large language models (LLMs), appear increasingly capable of performing sophisticated language-based tasks. These models can draft contracts, compose poetry, suggest travel itineraries, and even assist in generating text for government policy documents. For example, the UK’s AI tool "Redbox" has been used to streamline ministerial workflows. Around the world, governments are exploring whether AI could support policymaking; a 2024 report \citep{AIingov2024} revealed that over a third (37\%) of UK government departments actively use AI, with another 37\% either piloting (25\%) or planning (11\%) AI initiatives.

This interest is largely driven by the complex, interconnected global challenges policymakers face—challenges that are often difficult to address without relevant datasets, scientific insights, and scenario-analysis tools. Climate change, for instance, is an area where policymakers depend on computer simulations to predict future scenarios and shape policy, as evidenced by two decades of climate forecasting from the Intergovernmental Panel on Climate Change (IPCC). Such simulations involve complex calculations that surpass human cognitive abilities, similar to those used during the COVID-19 pandemic to model the virus's exponential growth and inform public health policies. These computational tools augment human understanding, helping us envision the consequences of our choices and responsibly shape the future.

While simulations have historically been designed by scientists, AI is now expanding these capabilities and introducing new applications to the policy cycle \citep{lv2022artificial}. AI can analyze vast amounts of social media data to capture public sentiment\footnote{https://radio.unglobalpulse.net/uganda/}, simulate policy instruments and their potential impacts \citep{rudd2024crafting}, identify valuable research \citep{susnjak2024automating}, optimize urban planning \citep{son2023algorithmic}, and even evaluate policies in real-time \citep{wirjo2002artificial}.

However, policymaking necessitates more than simulations or accurate predictions based on past data \citep{van2019bias}. While prediction—e.g. using AI to foresee probable scenarios based on historical trends—plays a role, it cannot address the full spectrum of policy challenges. 
 Policymakers instead must engage in responsible foresight, exploring a range of possible futures, including those that are unexpected, unintended and desirable. Responsible foresight \citep{uruena2021foresight} will thus require technical tools like AI to close the decision loop, but also a deep understanding of the interconnected social, economic, and environmental systems that influence future outcomes, alongside a commitment to ethical and sustainable decision-making. 


This paper examines the role of human-centric AI in supporting responsible foresight in policymaking. We call this field \emph{responsible computational foresight}. We argue that while AI can be instrumental in helping policymakers anticipate risks and seize opportunities, it should not replace, but rather augment, the ethical, social, and creative dimensions of human judgment.  AI should be viewed as a tool to enhance policymakers’ and societies' capacity for thoughtful future design, empowering them to navigate uncertainty and create strategies that are effective, sustainable, and ethically sound. Our goal is to foster an informed dialogue around the opportunities and risks associated with AI in policymaking. By doing so, we aim to contribute to the development of a dynamic, evolving framework that addresses the complex socio-technical dilemmas and ethical challenges inherent in algorithmic decision-making for public governance. This approach will help policymakers navigate the intricacies of these tools, ultimately aiming to enhance public welfare while safeguarding trust and accountability.

\section{Responsible foresight}

In an era of rapid technological advancements, global interconnectivity, and complex societal challenges, governments must increasingly anticipate future trends and uncertainties to craft effective, sustainable policies. This is where the practice of foresight becomes essential. Foresight refers to the systematic exploration of possible futures \citep{martin1995foresight}, enabling policymakers, futurists and foresighters to proactively shape strategies instead of merely reacting to emerging issues \citep{van2019bias}. However, the process of foresight must also be conducted responsibly to ensure that the actions we take today align with ethical principles and sustainability goals for future generations.

\textcolor{black}{
Responsible foresight can be defined as a structured and ethically-driven approach to exploring and shaping possible futures, grounded in methodological rigor that leverages high-quality data, scientifically sound analyses, and flexibility. But also, importantly,  incorporating core values like sustainability, equity, and inclusivity, helping policymakers anticipate risks, assess multiple scenarios, and make proactive, accountable decisions. This approach prioritizes empowering society to navigate uncertainty, fostering the creation of resilient and ethically-informed futures.}

\subsection{The role of responsible foresight in policymaking}

Responsible foresight goes beyond simply predicting or preparing for the future \citep{van2019toward}; it integrates values such as equity, environmental stewardship, and social justice into long-term thinking \citep{fuller2024responsible, bierwisch2024futures}, treating the future as an open space that we can actively shape through present choices. This approach enables policymakers, strategists, and leaders to not only anticipate opportunities and challenges but also to consider a broader range of outcomes and consequences that may be nonlinear or unexpected. By exploring multiple potential futures, responsible foresight supports better policymaking, allowing for more informed, flexible, and adaptive governance.

Rather than attempting to predict one definitive future, responsible foresight encourages the exploration of diverse scenarios, helping policymakers identify actions that can shape a preferable future. It serves as a guide for present decisions and actions, emphasizing proactive steps to prevent potential problems from arising in areas such as technology, social policy, and ethics.



\subsection{Key principles of responsible foresight in policymaking}

A responsible approach to foresight is not about accurate predictions but about empowering society to make responsible choices today. By presenting a range of potential futures, foresight enables innovation and emphasizes the ethical implications of decisions, ensuring that policies are crafted with both immediate and long-term goals in mind. This approach also values adaptability, recognizing the need for policies to evolve in response to shifting circumstances.
Long-term thinking through responsible foresight is essential for addressing complex global challenges like climate change, technological disruption, and socioeconomic inequality. While the future remains uncertain, responsible foresight provides a roadmap to navigate these complexities, equipping policymakers to build more inclusive, sustainable, and resilient futures.

\textcolor{black}{
In our work, we have identified several principles that are key to responsible foresight, which extend beyond general foresight practices to include fundamental aspects such as scientific rigor and data integrity. These principles are essential to ensuring that any computational tools used in responsible foresight uphold high standards and contribute to effective, reliable decision-making. Below are some of the core principles that are key to responsible foresight:}
\textcolor{black}{
\begin{itemize}
\item Sustainability and justice
\begin{itemize}
    \item \textbf{Sustainability}: Foresight should prioritize long-term environmental and societal sustainability to ensure future generations inherit a resilient and livable world.
\item \textbf{Equity and intergenerational justice}:  Responsible foresight should consider fairness not only among current populations but also between present and future generations. Policies should aim to prevent inequalities and ensure that future generations are not burdened by today’s decisions, balancing current needs with long-term impacts.
\item \textbf{Precautionary principle}: Minimize risks and prevent harm by carefully evaluating potential dangers, particularly in the face of uncertainty and emerging technologies.
\end{itemize}
\item Ethics, inclusion and transparency
\begin{itemize}
\item \textbf{Ethical considerations}: Evaluate the moral implications of future scenarios to ensure decisions avoid harming vulnerable populations or promoting unjust practices.
\item \textbf{Inclusivity and participation}: Actively involve diverse stakeholders in shaping visions of the future, including marginalized groups. This mitigates bias and ensures that decisions are grounded in a broad range of perspectives.
\item \textbf{Empowerment and capacity-building}: Enable citizens and organizations to engage proactively with foresight, fostering resilience and adaptability amid uncertainty.
 \item \textbf{Accountability and transparency}: Governments must remain accountable for the future they help shape, ensuring that foresight processes are transparent and open to public scrutiny.
\end{itemize}
\item Integrated systems and resilience
\begin{itemize}
\item \textbf{Systems thinking}: Consider the interdependence of economic, environmental, technological, and social systems to understand the ripple effects of decisions.
\item \textbf{Adaptability and responsiveness}: Responsible foresight requires policies to be designed with the flexibility to respond to unforeseen developments. Given the dynamic nature of global challenges, policies must be adaptable to changing data, emerging trends, and unanticipated risks, allowing for iterative improvements based on new information.
\end{itemize}
\item Iterative and exploratory practices
\begin{itemize}
\item \textbf{Exploration of multiple futures}: Rather than fixing on a singular, predicted future, responsible foresight explores multiple potential futures. This approach provides insight into how present-day actions could lead to various outcomes, empowering decision-makers with more nuanced options.
\item \textbf{Continuous monitoring and feedback loops}: As conditions evolve, foresight practices should include regular monitoring and feedback mechanisms to assess the impact of decisions and adjust course when necessary. This principle supports accountability by ensuring that foresight is not a one-time activity but an ongoing process that responds to new data and changing conditions.
\end{itemize}
\item Scientific rigor and data integrity
\begin{itemize}
\item \textbf{Scientific rigor}: Foresight must be underpinned by scientifically sound methodologies, ensuring that predictions and scenario analyses are based on reliable, validated approaches. This involves using rigorous models, transparent assumptions, and systematic methods to produce credible and robust insights.
\item \textbf{Data integrity}: The accuracy, quality, and transparency of the data used in foresight processes are paramount. Foresight tools should prioritize high-quality, unbiased data sources, ensuring that information is accurate, comprehensive, and up-to-date. Ensuring data integrity helps prevent the distortion of outcomes and maintains trust in foresight results.
\end{itemize}
\end{itemize}
}

\section{Computational modelling for responsible foresight}

While the future remains uncertain, modern civilization is structured around efforts to anticipate what lies ahead and much of that work is done through modelling approaches, whether in the short term (e.g. weather forecasts and traffic predictions) or over decades (such as climate projections and demographic trends). These forecasts help societies prepare for what may come, and as the pace and complexity of change seem to accelerate, so too has the need for effective anticipation.

Incorporating responsible foresight into policymaking increasingly will involve the use of algorithms to analyze complex data, predict potential outcomes, and offer insight into a range of possible futures. However, algorithms must be carefully designed to align with the key principles of responsible foresight listed before, ensuring they are not only scientifically rigorous but also ethically grounded, adaptable, and inclusive. By embedding these principles into algorithmic design and deployment, policymakers can leverage computational tools to support long-term, sustainable, and equitable decision-making.

\subsection{The rise of algorithms in policymaking}

\begin{figure}[ht!]
\centering
\includegraphics[width=0.7\textwidth]{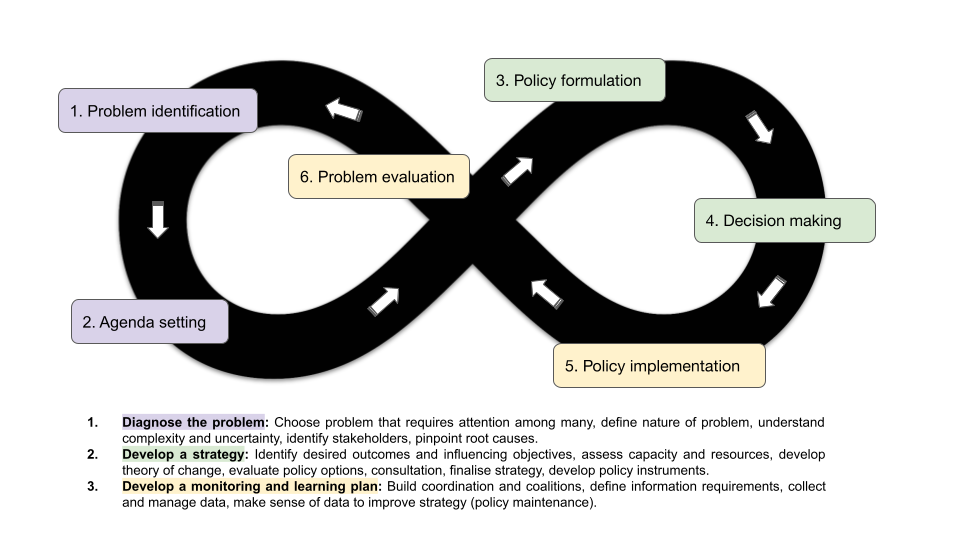}
\caption{\label{fig:cycle}Representation of the cyclical nature of the policymaking cycle.}
\end{figure}

Public policymaking is inherently a cyclical and iterative process, encompassing stages such as the identification of societal needs, formulation of agendas and policy alternatives, adoption of policy measures, real-world implementation, and finally, evaluation of outcomes, followed by refinements and improvements \citep{jann2017theories}. This cycle, as illustrated in \figurename{ \ref{fig:cycle}}, represents a structured pathway that allows for continuous learning and adaptation. Yet, in practice, policymakers are also subject to external influences beyond this structured cycle, including budget constraints, public opinion, the interests of civil society, ideological stances, media narratives, economic conditions, and ongoing scientific research—all of which shape and constrain policy decisions in significant ways.


When analyzing each step of the policymaking cycle, it becomes clear that algorithmic and computational tools have the potential to enhance decision-making at every phase. Numerous prototypes and small-scale proofs of concept already demonstrate how AI and algorithmic methods can support critical policymaking functions. 
For example, at the initial stage of identifying needs, AI applications have proven instrumental in rapidly synthesizing large datasets and detecting emerging patterns in real time, enabling more agile responses. This capability is particularly crucial in the context of today’s "infodemic", where over 90\% of online data has been generated in recent years, with approximately 80\% of this data being unstructured and thus challenging to process without sophisticated tools.
In this sense, the Victoria State Government in Australia implemented a “syndromic surveillance” program, which monitors reported symptoms and patient characteristics across hospitals. Within just four months, this tool allowed policymakers to identify and address six distinct public health issues. Similarly, the United Nations Global Pulse initiative in Uganda utilized a radio content analysis tool to capture public concerns, which informed the planning and prioritization of development projects. These examples highlight AI’s potential to aggregate and analyze complex data, providing actionable insights that can guide policy decisions at scale and speed previously unattainable.

Despite this potential, the far-reaching impact of policy decisions mandates careful scrutiny of these technologies, as they bring inherent risks and ethical concerns alongside their benefits. The public’s trust in government is closely tied to transparency, fairness, and accountability in decision-making. Accordingly, any deployment of AI in policymaking must be underpinned by robust frameworks that ensure these technologies are used responsibly. Principles of fairness, transparency, and accountability must be integral from the outset, ensuring that AI-driven decisions are scientifically valid and ethically sound. Before policymakers commit to these technological tools, it is essential to establish a foundation that prioritizes these principles.


\subsection{Sub-fields of responsible computational foresight}

Our efforts to glimpse the future fall into two primary approaches: prediction and foresight. Prediction is grounded in understanding a system’s components and how they interact, allowing us to estimate how the system’s state might evolve in response to various changes. With certain systems, we can make reasonable predictions that yield one likely outcome. However, in most cases, futures research recognizes that the future cannot be predicted with complete accuracy. Instead, it’s a matter of probabilistic thinking. Often, a range of distinct outcomes is possible, requiring us to rely on foresight \citep{mcmaster1996foresight}. Foresight combines methods for gathering and assessing information relevant to multiple potential outcomes, evaluating their implications, and discussing which scenarios may be most desirable. For issues like climate change, geopolitical shifts, or evolving lifestyle patterns, foresight emphasizes exploring the full range of possible and sometimes unexpected futures, not merely those that seem probable. This process may be organized from the top down (as in strategic foresight \citep{brandtner2021artificial}) or through participatory approaches \citep{van2019bias}, involving diverse voices and perspectives. 
We review now a series of tools aimed at responsible foresight (some of the tools taking inspiration from the ones listed in \cite{GESDA2024}), and discuss the role of computational modeling in each of those, presenting recent and novel examples from the literature.

\subsubsection{Superforecasting and prediction markets}
While anticipating the future is challenging, it is possible to achieve valuable insights, as demonstrated by approaches like superforecasting and prediction markets. These methods synthesize various data sources to produce meaningful forecasts. Superforecasting \citep{tetlock2016superforecasting}, a skill typically honed by individuals (but extendable also to computational methods/agents) known for their consistent accuracy in making probabilistic predictions, demonstrates that certain methods and cognitive strategies can enhance the ability to forecast uncertain events (e.g. disciplined reasoning, openness to updating beliefs, and careful consideration of probabilities). Prediction markets \citep{wolfers2004prediction}, on the other hand, harness the collective intelligence of groups, allowing individuals to buy and sell shares in future outcomes. These markets aggregate diverse perspectives, synthesizing information in a way that often produces highly accurate forecasts. Both superforecasting and prediction markets represent powerful tools for approaching future uncertainties, particularly in complex environments where multiple factors intersect. 

Forecasting specifically has seen significant advances through AI, which has excelled for over a decade in well-defined tasks like weather prediction. Recently, however, new advanced generative models geared toward superforecasting complex, world-changing events—across domains like society, economy, politics, and science—are emerging as new tools \citep{zou2022forecasting, karger2024forecastbench,pratt2024can}. Specifically, large language models (LLMs) trained on past human forecasts are beginning to achieve performance levels that approach, though do not yet surpass, human forecasters. In fact, “crowds” of LLMs have sometimes reached the accuracy of human superforecasters \citep{schoenegger2024wisdom}.

A seemingly promising development is the use of LLMs as forecasting assistants: recent studies show that human superforecasting accuracy improves by 23\% when LLMs are used as supportive tools, without diminishing forecast diversity \citep{schoenegger2024ai}. These models are also beneficial in refining prediction market data by aggregating and weighing various predictions to produce more accurate forecasts \citep{gruen2023machine}. However, some studies suggest more conservative expectations \citep{abolghasemi2024humans}, highlighting the need for caution when deploying these as assistants. Ultimately, advanced models such as LLMs are starting to show potential in enhancing human forecasting capabilities and as a tool for refining collective predictions in complex domains.

\textcolor{black}{There is however a critical risk that the use of superforecasting may inadvertently act as a self-fulfilling prophecy. Predictions generated or amplified by this tool of foresight could influence decision-makers, markets, or public sentiment, thereby increasing the likelihood of the forecasted outcomes occurring simply because they were predicted. This underscores the importance of the rest of the tools presented in this section, which aim to empower humans to imagine and design the future, using predictions not as fixed outcomes but as guides to inform the actions needed to shape more desirable and equitable possibilities.
}


\subsubsection{World simulation, surrogate modelling or emulation}
Beyond probabilistic forecasting tools, some researchers and practitioners have been working toward a more holistic form of anticipation through world simulation—advanced modeling frameworks that strive to capture the vast, interconnected nature of global systems. These simulations aim to model key physical, economic, social, and ecological processes, offering a dynamic platform where different scenarios can be tested to observe potential outcomes. These tools are essential in addressing the major challenges humanity faces in the 21st century.  World simulation allows for the exploration of various "what if" scenarios and provides policymakers and stakeholders with a visualized, interactive way to grasp the consequences of specific decisions. For instance, simulations can illuminate the long-term impact of climate policy, the ripple effects of geopolitical shifts, or how different energy policies might influence socioeconomic conditions across regions. By constructing a high-fidelity representation of interconnected systems, world simulation helps translate complex data into a form that supports responsible foresight.

World simulation is not a new concept; it traces back to the Limits to Growth report in 1972, which used early computer models to examine the potential consequences of exponential economic and population growth within a finite resource environment. Today, the concept has evolved through the advent of digital twins \citep{lv2022artificial}—digital replicas of physical objects, systems, or processes contextualized within their digital environments. Digital twins enable organizations to simulate real-world situations and potential outcomes, helping them make better-informed decisions. For example, integrated assessment models \citep{parson1997integrated} are a form of digital twin used by scientists and policymakers to support decision-making by modeling the intricate interactions among society, the environment, and the economy.

Recently, machine learning has further enhanced these models’ predictive capabilities by capturing underlying dynamics with greater precision. For example, AI shows promise in emulating complex biological systems \citep{stolfi2021emulating}, while hybrid models that combine machine learning with physics-based simulations have begun to yield significant predictive improvements \citep{willard2020integrating}. Although many challenges remain in creating digital twins that can fully replicate real-world systems, promising early results in AI-based world simulation are beginning to emerge \citep{yang2023learning}, even as accurate proxies of certain human behaviours \citep{park2023generative}.


\subsubsection{Simulation intelligence}


Simulation intelligence is an emerging field that combines advanced simulations and AI to analyze and understand complex systems \citep{lavin2021simulation}. By integrating AI with high-fidelity simulations (e.g. as the previous subsection describes), simulation intelligence enables the exploration of a complex system, "what-if" scenarios within it and the discovery of optimal control policies. This approach could empower organizations and policymakers to comprehend and manage complex systems, by employing computational tools to explore potential system interventions. This can be used as a tool for discovery, enabling a "closed generative loop" that can generate new insights for applications like drug discovery or crafting effective policies in urban planning \citep{son2023algorithmic} and climate \citep{rudd2024crafting}.
See for example \cite{stock2024plant}, which explores the transformative potential for the plant science community of the simulation intelligence motifs, from understanding molecular plant processes to optimizing greenhouse control, arguing that these can potentially revolutionize breeding and precision farming towards more sustainable food production. Similarly, simulation intelligence can be used for generating new policy insights that balance sustainability trade-offs \citep{rudd2024crafting}. 

Simulation intelligence involves a diverse set of techniques, including simulation-based inference, causal modeling, agent-based modeling, and probabilistic methods, among others. Coordinating these approaches offers significant potential to advance scientific discovery and inform policymaking. This could range from solving inverse problems in fields like synthetic biology and climate science to guiding nuclear energy experiments and predicting complex behavior in socioeconomic systems \citep{lavin2021simulation}. The scope of simulation intelligence is extensive, especially as emerging research integrates different data modalities for enhanced outcomes—for example, combining natural language processing with simulation intelligence by using large language models (LLMs) alongside simulation and optimization techniques to tackle complex challenges \citep{rasal2024optimal}.

\subsubsection{Scenario building and narrative-based techniques}

Scenario-building and foresight techniques, which use narrative and visual tools to aid in anticipation, are crucial, supporting this anticipatory process by presenting a structured way to explore multiple potential futures \citep{drew2006building}. Scenario-building involves creating plausible, narrative-driven models of the future, each based on a different set of assumptions and drivers. By examining how factors such as technological advances, social trends, or regulatory shifts might unfold, scenarios help humans identify a range of possible futures rather than one fixed outcome. This approach enables decision-makers to consider the broader implications of their choices, evaluate diverse pathways, and recognize which scenarios align with their values and goals. When integrated with robust foresight methods, scenarios can illuminate both risks and opportunities, empowering leaders to make more informed and adaptable decisions.

Advanced AI systems, such as LLMs, are now employed in scenario generation to explore unlikely or edge cases that could advance Industry 5.0 \citep{chang2024llmscenario}. Beyond industry, these models could also be used to examine unexpected outcomes resulting from policy interventions. LLMs additionally contribute by generating narratives—vivid glimpses into potential futures that help us envision life within them \citep{zhao2023more}. AI-assisted scenario building has shown promise in creating a diverse range of scenarios, offering valuable insights that can support progress toward sustainable development \citep{carlsenai}.




\subsubsection{Participatory futures and futures literacy}

Apart from scenario-building, participatory futures and futures literacy initiatives further help communities and organizations to think critically about the future, focusing on making it a more inclusive and engaged process. This adds another crucial dimension to responsible foresight by actively involving communities and stakeholders in the foresight process. Participatory futures \citep{gidley2009participatory} seek to democratize future-planning by engaging people from diverse backgrounds, including those often underrepresented in policy discussions, to share their insights, hopes, and concerns for the future. This approach recognizes that incorporating a wide array of perspectives can mitigate bias and help ensure that proposed strategies are socially inclusive, innovative and resilient. Futures literacy, meanwhile, involves developing the skills, datasets and frameworks that allow individuals and new technologies to support thinking critically and creatively about the future. It equips people and machines to understand the purpose and limitations of foresight exercises, enhancing their capacity to interpret, question, and act on future-oriented insights. Together, participatory futures and futures literacy contribute to responsible foresight by fostering a collective sense of ownership over the future, empowering communities to envision and shape futures that reflect their values. \textcolor{black}{By doing so, AI can contribute to education for democracy, fostering the design of participatory futures, where diverse voices shape tomorrow's decisions. This approach empowers individuals and communities to actively engage in envisioning and co-creating equitable and inclusive futures, ensuring that the democratic process extends into shaping the world of tomorrow.}

Envisioning future scenarios based on events and (in)actions is essential for democratic engagement, yet it is often the domain of experts with deep insights into social, political, environmental, or technological trends. AI co-writing tools \citep{tost2024futuring} have shown promise in reducing barriers to participation, empowering non-experts to imagine future scenarios and craft fictional stories set in speculative futures. LLMs extend scenario-based design by generating "value scenarios" \citep{jung2023toward}, which foster critical, systemic, and long-term thinking in design practices, technology development, and deployment. Some of these models are also applied in participatory planning, such as generating urban plans that integrate the diverse needs of residents \citep{zhou2024large}. AI holds the potential to enhance democratic and participatory processes by facilitating more accessible methods of preference elicitation and simulating collective preferences \citep{gudino2024large, feffer2023preference}. By simplifying how preferences are gathered, AI can broaden public participation, making it easier for individuals to express their views. Additionally, AI can augment existing citizens assemblies \citep{mckinney2024integrating} or data on political preferences \citep{feffer2023preference}, providing deeper insights into public opinion and supporting more responsive, inclusive decision-making. Although few works exist at the intersection of futures literacy and AI, emerging studies illustrate how futures literacy can help envision the future of a world with AI and its broader impacts \citep{liveley2022ai,leander2020critical}.

\textcolor{black}{\subsubsection{Hybrid intelligence and human-computer interaction}}

\textcolor{black}{
While not very developed as a field yet, the topic of hybrid human-artificial intelligence \citep{schlobach2022hhai2022} will play a pivotal role in advancing responsible foresight by combining the complementary strengths of both human and machine intelligence. 
It is important to note that AI represents a fundamentally different kind of intelligence than that of humans, one that processes data rapidly and accurately, operates without fatigue or distraction, and performs complex computations that can optimize systems with efficiency. While these capabilities offer clear advantages to our human mind, AI’s limitations—such as the current absence of common sense and reasoning—raise concerns about its reliability in nuanced decision-making. Without a nuanced understanding of context and reliable data (e.g. often unavailable for black swan events), AI can unintentionally perpetuate biases or preconceptions present in its training data, creating unintended consequences if left unchecked. This is further compounded by AI’s need for vast amounts of data and high energy demands, highlighting an important challenge for responsible foresight: developing ethical, sustainable, and context-sensitive AI systems that fully support human intelligence in future decision-making.}

\textcolor{black}{
Human intelligence, by contrast, is adaptive and integrative. Humans generalize quickly from few examples, adapting to novel contexts and switching fluidly between immediate needs and long-term goals. Our capacities for imagination, empathy, and ethical judgment make us particularly well-suited for responsible foresight. When envisioning future scenarios, humans consider not just logical outcomes but also social, emotional, and ethical dimensions. This integrative perspective remains essential in tackling complex, open-ended challenges like climate change or global inequality, which require foresight to consider both anticipated and unforeseen impacts on society.}

\textcolor{black}{
Effective human-computer interaction (HCI) is key to this collaboration between humans and machines, ensuring that AI systems are designed to be intuitive, transparent, and responsive to human needs. Interactive interfaces, explainable AI models, and participatory design processes can facilitate a seamless exchange of insights between humans and machines, enabling users to critically engage with AI outputs and refine them based on their expertise and ethical considerations. Importantly, this interaction must address the balance between foretell and forsay: humans should retain the ability to question, contradict, or reshape the foresights provided by AI, preventing these outputs from being treated as unchallengeable truths or prophetic decrees. By embedding mechanisms for contestation and oversight, such as tools for scrutinizing input data, questioning assumptions, and modeling alternative scenarios, humans can maintain control and avoid the risk of "Garbage In, Gospel Out". This ensures that foresight remains a deliberative process, grounded in human judgment and ethical reflection, rather than a deterministic one dictated by machines. By fostering interactive partnerships, hybrid intelligence enables the co-creation of futures that are not only informed by data but also guided by shared values, ethical considerations, and diverse perspectives.
}








\subsection{An integrative framework of responsible computational foresight}

These complementary methods—Superforecasting, prediction markets, world simulation, simulation intelligence, scenario-building, participatory futures, futures literacy and hybrid intelligence—form a powerful toolkit for responsible computational foresight. Each approach contributes unique insights: Superforecasting harnesses the wisdom of skilled forecasters to generate highly accurate predictions; prediction markets leverage collective knowledge; world simulation creates virtual environments to capture the complexity of our social, economic and environmental systems; and simulation intelligence uses AI to design control strategies within simulation worlds providing meaningful insights on efficient and resilient pathways for the future. Scenario-building further supports human imagination in identifying a range of possible futures. Participatory futures invite diverse stakeholders into the process, democratizing foresight by integrating varied perspectives, while futures literacy builds a capacity for flexible, long-term thinking. \textcolor{black}{Hybrid intelligence and human-computer interaction add another vital layer, bridging the computational and human dimensions of foresight by enabling collaboration, contestation, and creativity. Together, they ensure that these methods not only generate robust insights but also remain aligned with human values, ethics, and agency.} Various tools already embody the principles of responsible computational foresight, such as those designed for AI-assisted deliberation \citep{zhang2023deliberating, lyu2023design} or for developing action plans in disaster scenarios \citep{goecks2023disasterresponsegpt}.

When applied thoughtfully, these tools strengthen our ability to explore a spectrum of possible futures, reinforcing ethical decision-making and enhancing resilience in the face of uncertainty. Rather than aiming for precise predictions, this toolkit enables us to anticipate and prepare for a range of outcomes, helping society to navigate complexity with informed, adaptive, and inclusive strategies. By doing so, responsible computational foresight not only supports current decision-making but also empowers us to shape sustainable and equitable paths forward for generations to come.

\begin{figure}[ht!]
\centering
\includegraphics[width=0.9\textwidth]{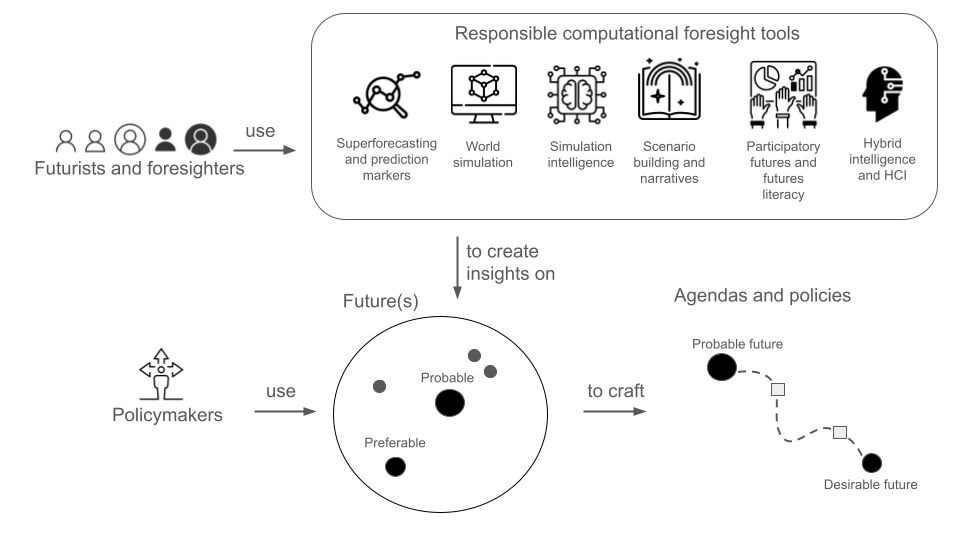}
\caption{\label{fig:foresights}Diagram illustrating the flow of responsible computational foresight. Futurists and foresighters employ responsible foresight tools to generate a diverse array of insights about the future—covering risks, potential consequences, likely outcomes, and preferable pathways, among others. Policymakers (who may also act as futurists or foresighters) then leverage these insights to inform decisions, shape agendas, and develop policy trajectories that steer society toward more desirable futures.}
\end{figure}


\figurename{ \ref{fig:foresights}} illustrates the process outlined in this position paper, showing how policymakers are empowered by a range of computational tools that support responsible foresight. These tools enable forecasting, complex systems simulation, creative scenario-building, and the integration of diverse perspectives, enhancing the policymaking process by providing a more comprehensive understanding of future possibilities. By using these methods, policymakers can better diagnose challenges, identify root causes, and establish an intentional, forward-looking agenda. This approach aids in selecting policies that effectively support targeted change and helps to determine optimal courses of action, all while incorporating real-time insights through augmented democratic participation \citep{gudino2024large}.
In addition to these core methods, other related fields, such as computational diplomacy \citep{cederman2023computational}, play an essential role in planning agendas and policies towards more desirable futures. Here, tools like game theory assist policymakers in crafting effective strategies within the complex landscape of global, multicultural interests. This holistic, computationally augmented approach to foresight positions policymakers to address both current and future challenges in a more inclusive, adaptive, and resilient way.





Responsible computational foresight is about supporting humans in understanding and designing the future. 
This is in line with the tenets of the field of future design, citing \citep{miller2011opinion}:
\emph{“We should abandon the effort to try to be so clever that we can choose the right model, find the right data, or make the best guess. There is no way to outsmart the complexity of reality; unforeseeable novelty is a certainty. Instead, the approach should be to try and develop the capacity to use the future in a range of different ways, and not be limited by prediction or by narrow conceptions of a desired future. It is about being Futures Literate.”.} 

\section{Discussion}

This work establishes key principles for responsible computational foresight and introduces a suite of tools that are actively in use, many of which are actively being shaped by AI—from superforecasting and world simulation to participatory futures.

For responsible foresight, AI’s role should center on augmenting human intelligence rather than replacing it. AI should enhance human foresight by illuminating patterns and possibilities that might otherwise go unnoticed. Instead of attempting to replicate human decision-making, the focus of responsible foresight is on leveraging AI’s strengths—data processing, pattern recognition, and scenario analysis—to support and expand human thinking.

This augmentation approach is crucial because while AI excels in structured, well-defined environments, it is often limited in open-ended settings where social, political, and ethical complexities resist quantification. Games like Go illustrate AI’s capabilities within a rule-bound system, but translating these abilities to real-world foresight, especially in policy design, is another matter entirely. Responsible foresight goes beyond optimizing for efficiency; it requires insight into human values, a consideration of diverse stakeholder perspectives, and the ability to adapt to shifting social landscapes—all elements that are difficult to formalize for AI.

Human decision-making in responsible foresight includes assessing potential social impacts, navigating political dynamics, and making ethically informed choices. It demands critical and contextual intelligence: questioning assumptions, building consensus, evaluating evidence, and applying ethical judgment—all facets that AI alone currently cannot provide. Nevertheless, AI’s ability to analyze extensive datasets, model complex systems, and simulate alternative futures presents a promising pathway to support responsible foresight, enabling policymakers to gain insights that extend beyond what human cognition alone can offer.

For instance, AI’s capacity to simulate “digital twins”—virtual models of physical or social systems—can aid foresight by allowing policymakers to explore various scenarios, assess risks, and anticipate consequences before implementing policies. As a long-term strategy, AI could help dismantle traditional silos in policymaking by integrating data across domains such as health, education, and labor, providing a more comprehensive view of complex issues. By expanding the range of information and perspectives available, AI has the potential to support more cohesive, informed, and forward-looking policy frameworks.

To realize this potential, AI in responsible foresight must be conceived as an assistive tool—a cognitive exoskeleton that enables policymakers to navigate complexity, envision a range of desirable futures, and critically assess the impacts of various decisions. This role calls for foresight-focused AI designed to identify patterns, simulate plausible outcomes, and help policymakers weigh risks in alignment with ethical principles.

Ultimately, the goal of responsible computational foresight is to create a partnership where AI’s computational power complements human judgment and ethical insight. By fostering such a collaboration, policymakers and communities can harness AI’s strengths to meet the challenges of tomorrow, navigating towards a future that is ethically grounded, resilient, and responsive to the needs of society. Imagine the potential if AI were purposefully developed to align with and enhance human foresight, empowering us to anticipate and address the socio-environmental challenges that will shape the future of humanity.

\bibliography{paper-refs}

\end{document}